# Piecewise linear topology, evolutionary algorithms, and optimization problems

Andrew Clark

andrew.clark@thomsonreuters.com

Abstract

Schemata theory, Markov chains, and statistical mechanics have been used to explain how evolutionary algorithms (EAs) work. Incremental success has been achieved with all of these methods, but each has been stymied by limitations related to its less-than-global view. We show that moving the question into topological space helps in the understanding of why EAs work.

**Purpose.** In this paper we use piecewise linear topology to attempt a global explanation of why EAs work. We show that when an EA solves a convex optimization problem, piecewise linear-mappings (PL-mappings) are the simplicial mappings of the original polytopes. We also demonstrate that, since a quotient space exists, the union of the simplicial vertices generates the convex hull of the convex optimization problem the EA is seeking to solve. The unique convex hull is found via tight triangulations, which are a form of combinatorial topology.

**Methods**. Piecewise linear topology, basic optimization theory, Borel algebra, algebraic geometry, and tight triangulations are used to study why EAs work.

 **Results.** It is demonstrated that the use of piecewise linear topology in a combinatorial form helps explain why EAs work**.**

**Conclusions.** Moving the "why EAs work" question into topological space is found to be helpful in understanding this basic EA question. The basic topological conditions are married to a Borel algebra in order to account for EA operators such as mutation and selection, and combinatorial topology demonstrates the EAs' ability to find the unique convex hull.

**Piecewise linear topology**

The following is a sketch of the necessary piecewise linear topology (PL-topology). It follows Sanderson and Rourke [1] as well as Hudson [2].
- A space homeomorphic to a polyhedron is called a topological polyhedron. This space, often referred to as a t-polyhedron, contains the most important objects of finite dimensional topology, such as smooth manifolds.
- In PL-topology there are four categories: *F*, *P*, *K,* and *A*. The objects of *F* are the t-polyhedra and their morphisms' continuous mappings. *P* are the polyhedra, and their morphisms are the PL-mappings. PL-mappings linearly transform the convex polytopes of some covering of the domain into polytopes of some covering of the range. The objects of *K* are simplicial mappings, i.e., PL-mappings that linearly transform each simplex of the domain onto the mappings of the range. Finally, *A* consists of the abstract complexes (a-complexes) and their simplicial mappings.

- An a-complex in *A* is at most a countable set of $\overline{A}$ with a system of finite subsets called simplicies, which satisfy the conditions such that for each simplex *σ* the system also contains all subsets of the simplex or faces of *σ*, and each simplex is a face of at most a finite number of other simplicies.
- There are funct*ors* in PL-topology:

$$\Im \xleftarrow{t} P \xleftarrow{p} K \xrightarrow{a} A$$

- Since the polyhedron defines a topological space, and since PL-mappings are continuous, *t:t(P)* is the space of the polyhedron *P*. Since each complex defines a polyhedron and a simplicial mapping of the complexes of the PL-mappings is denoted as *P:p(K,)*, the body or skeleton of the complex is denoted as $|K|$.
- The set of vertices of *K* contains subsets. These are the sets of the vertices of the simplicies in *K*, which define the a-complexes, and the simplicial mappings of complexes, which define simplicial mappings of the corresponding a-complexes. *a* and *a(K)* are called the scheme of complex *K*.
- The functors do not have natural inverses. However, they do have equivalences, if there are suitable quotient categories. The associated isomorphisms, given the existence of the quotient categories, are: homeomorphisms in $\Im$, the PL-homeomorphisms in *P*, and the simplicial isomorphisms in both *K* and *A*.
- Any polyhedron *P* that is a skeleton of some complex *K* in which *K* is a known triangulation of *P* means the scheme of *K* is called an abstract triangulation of *P*. Also, given a PL-mapping $f : P \rightarrow Q$, there exist triangulations *K* for *P* and *L* for *Q*. *f*, therefore, is a simplicial mapping of *K* into *L*.
- A PL-structure is defined by the homeomorphism $\tau : T \rightarrow P$ of a t-polyhedron. A PL-structure implies two homeomorphisms: $\tau_1 : T \rightarrow P_1$ and $\tau_2 : T \rightarrow P_2$. The two homeomorphisms are considered the same structure if $\tau_1 \tau_2^{-1}$ is a PL-homeomorphism. These two homeomorphisms define an equivalent structure if *P₁* and *P₂* are also PL-homeomorphic.
- A t-polyhedron with a fixed PL-structure is also a polyhedron.

- The relation of the combinatorial equivalence in *K* implies, via the functor *a*, a new equivalence relation in *A*. In order to formulate this relation in *A*, we need to define the operation of stellar subdivision.
- Stellar subdivision is the join of two simplicies, $\sigma_1^{n_1}$ and $\sigma_2^{n_2}$, whose vertices are in general positioned in vector space *R* and are defined as their convex hull.

**Evolutionary algorithms (EAs)**

A generic EA (or genetic algorithm [GA]) assumes a discrete search space *H* and a function

$$f : H \to \mathbb{R},$$

where *H* is a subset of the Euclidean space $\mathfrak{R}$.

The general problem is to find

$$\arg \min_{X \in H} f,$$

where *X* is a vector of the decision variables and *f* is the objective function.

With EAs it is customary to distinguish *genotype*–the encoded representation of the variables–from *phenotype*–the set of variables themselves. The vector *X* is represented by a string (or chromosome) *s* of length *l* made up of symbols drawn from an alphabet *A* using the mapping

$$c : A^l \to H.$$

If the domain of *c* is total, i.e., the domain of *c* is all of $A^l$, *c* is called a decoding function. The mapping of *c* is not necessarily surjective. The range of *c* determines the subset of $A^l$ available for exploration by an EA.

The range of *c*, $\Xi$

$$\Xi \subseteq A^l$$

is needed to account for the fact that some strings in the image $A^l$ under *c* may represent invalid solutions to the original problem.

The search space $\Xi$ can be determined by either Shannon or second-order Renyi entropy. If the decision variables *X* are independent, Shannon entropy applies. If the decision variables are correlated, then second-order Renyi entropy applies. A minimization of either entropy will define the feasible search space $\Xi$.

The string length *l* depends on the dimensions of both *H* and *A,* with the elements of the string corresponding to *genes* and the values to *alleles*. This statement of genes and alleles is often referred to as genotype-phenotype mapping.

Given the statements above, the optimization becomes:

$$\arg\min_{S \in L} g,$$

given the function

$$g(s) = f(c(s)).$$

Finally, with EAs it is helpful if *c* is a bijection. The important property of bijections as they apply to EAs is that bijections have an inverse, i.e., there is a unique vector *x* for every string and a unique string for each *x*.

## EAs, PL-topology, and optimization

Before linking EAs to optimization and the associated PL-topology, some basics of optimization theory are needed.

In optimization theory the inequalities or constraints that define the facets of the feasible set are characterized by a polyhedron. The feasible set is defined as nonempty and compact. In general, the feasible set is described as the intersection of the *m* closed half-spaces, i.e., convex polytopes. Convex programming problems are those for which the cost or objective function *f* is convex and *C*–the feasible set–is also convex. Convexity permeates all optimization problems, including those that are discrete.

Convexity's importance to optimization can be stated as follows:

- A convex function has no local minima that are not global minima.
- A convex set has a nonempty relative interior.
- A convex set is connected and has feasible directions at any point.
- A nonconvex function can be "convexified" while maintaining the optimality of its global minima.
- The existence of a global minimum of a convex function over a convex set is characterized in terms of the directions of recession.
- A polyhedral convex set is characterized in terms of a finite set of extreme points and extreme directions.
- A real-valued convex function is continuous and is differentiable.
- Closed convex cones are self-dual with respect to polarity.
- Convex, lower semi-continuous functions are self-dual with respect to conjugacy.

A convex polyhedral set is defined in the following manner:

$$conv(P) = \{v_1, v_2, ..., v_n\} + C,$$

i.e., there exists a convex hull [*conv(P)*], a nonempty finite set of vectors $\{v_1, v_2, ..., v_n\}$, and a finitely generated cone *C*.

To solve the convexity problem, optimization theory states that if there are global minima, there needs to exist a unique convex hull. If the hull exists, a simplex[a] can be formed and a solution generated. There are other ways of solving the convexity problem such as Nelder-Mead and others, but we let the simplex stand in for these other methodologies.

Can EAs find the unique convex hull if the global minima exist? The answer is yes. We start our proof by demonstrating the existence of equivalence classes in an EA environment. In both Radcliffe [3] and Radcliffe and Surry [4] a problem domain $f$ consists of a set of problem instances $A^I$, each of which takes the form of a search space (of candidate solutions) $H$, together with some fitness function defined on that search space (a fitness function will be defined below). A characterization of the domain specifies a set of equivalences among the solutions for any instance of $A^I$. These equivalences induce a representation made up of a representation space (of chromosomes, in EA terminology) and a growth function $c$ mapping chromosomes to the objects in $H$. A chromosome $s$ is a string of alleles that indicates that $s$ satisfies a particular equivalence on $H$.

Both papers postulate (and prove) a problem-dependent characterization that captures knowledge about the problem domain. This characterization mechanically generates a formal representation (a representation space and a growth function) for any instance of the problem by defining a number of equivalences over the search space. These equivalences induce subsets of the search space thought to contain solutions with related performance, possibly as partitions generated by equivalence relations or simply as groups of solutions sharing some characteristic. For a given solution the pattern of its membership in the specified subsets is used to define its alleles and possibly its genes. Although in some problems the search space can be partitioned orthogonally (meaning that all combinations of alleles represent valid solutions), this is not always the case.

To tie the work just discussed to PL-topology, remember that an equivalence relation is a relation that partitions a set so that every element of the set is a member of one and only one cell of the partition. A quotient space (or a quotient category) is defined to be the set of equivalence classes of elements of the topological space.

In PL-topology the objects of $F$ are the t-polyhedra, and their morphisms are continuous mappings. $P$ are the polyhedra, and their morphisms are the PL-mappings. PL-mappings linearly transform the convex polytopes of some covering of the domain into polytopes of some covering of the range (the instances referred to above). In optimization theory the inequalities or constraints that define the facets of the feasible set (the subsets of the search space $H$) are characterized by a polyhedron $P$. The feasible set is defined as nonempty and compact. In general, the feasible set is described as the intersection of the $m$ closed half-spaces, i.e., convex polytopes.

If each constraint in an optimization problem defines a half-space, the feasible set formed by this intersection of half-spaces is a simplex. Since $K$ is the simplicial mapping or the PL-mapping that linearly transforms each simplex of the domain onto the mappings of the range, the subsets of the vertices of *K* are the feasible sets generated by the EA.

The functors in PL-topology have equivalences, if there are suitable quotient categories. The associated isomorphisms, given the existence of the quotient categories, are homeomorphisms in $\Im$, the PL-homeomorphisms in *P,* and the simplicial isomorphisms in both *K* and *A*.

To show that a quotient space (or quotient category) exists for EAs, remember that each instance of the EA optimization problem defines a number of equivalences over the search space, i.e., a partitioning of the search space generated by equivalent relations. The set of equivalent classes generated by an EA is a quotient space.

To return to optimization theory, by definition the set of all points that can be expressed by a convex combination of extreme points (vertices) is called the convex hull of the given extreme points. By stellar subdivision the join of two simplicies, $\sigma_1^{n_1}$ and $\sigma_2^{n_2}$, whose vertices (extreme points) are in a general position in vector space *R* , is defined as their convex hull. Since EA generates subsets of the vertices *K*, EAs can also create the convex hull built on vertices. But can EAs find the unique convex hull that is the optimal solution occurring at the vertex of a simplex? This question will be answered below.

**EA operators and the search for the optimal solution**

The operators by which EAs search for the optimal solution are set out in the following statements from Coello *et al.* [5] (their notation is used with some slight modifications):

> Let *H* be a nonempty set (the individual or search space), $u^i_{i \in \mathbb{N}}$ a sequence in $\mathbb{Z}^+$ (the parent populations), $u^{'(i)}_{i \in \mathbb{N}}$ a sequence in $\mathbb{Z}^+$ (the offspring population sizes), $\phi: H \to \mathbb{R}$ a fitness function, $\iota: \cup_{i=1}^{\infty}(H^u)^{(i)} \to$ {true, false} (the termination criteria), $\chi \in$ {true, false}, *r* a sequence $r^{(i)}$ of recombination operators $r^{(i)}: X_r^{(i)} \to T(\Omega_r^{(i)}, T\ H^{u^{(i)}}, H^{u'^{(i)}})$, *m* a sequence of $\{m^{(i)}\}$ of mutation operators in $m^i$, $X_m^{(i)} \to T(\Omega_m^{(i)}, T\ H^{u^{(i)}}, H^{u'^{(i)}})$, *s* a sequence of $\{s^i\}$ selection operators $s^{(i)}: X_s^{(i)} \times T(H, \mathbb{R}) \to T(\Omega_s^{(i)}, T((H^{u^{(i)}+\chi\mu^{(i)}}), H^{\mu^{(i+1)}}))$, $\Theta_r^{(i)} \in X_r^{(i)}$ (the recombination parameters), $\Theta_m^{(i)} \in X_m^{(i)}$ (the mutation parameters), and $\Theta_s^{(i)} \in X_s^{(i)}$ (the selection parameters).

Coello *et al*. [5] define the collection $\mu$ (the number of individuals) via $H^\mu$. The population transforms (PTs) are denoted by $T: H^\mu \to H^\mu$, where $\mu \in \mathbb{N}$. However, some EA methods generate populations whose size is not equal to their predecessors'. In a more general framework $T: H^\mu \rightarrow H^{\mu'}$ can accommodate populations that contain the same or different individuals. This mapping has the ability to represent all population sizes, evolutionary operators (EOs), and parameters as sequences.

The execution of an EA typically begins by randomly sampling with replacement from $A^l$. The resulting collection is the initial population, denoted by *P(0)*. In general, a population is a collection $P = (a_1, a_2, ..., a_\mu)$ of individuals, where $a_i \in A^l$, and populations are treated as n-tuples of individuals. The number of individuals ($\mu$) is defined as the population size.

By treating populations as n-tuples we can define a convex cell *A* of $E^n$ (an n-dimensional Euclidean space, our PL-mapping space), where *A* is a compact, nonempty subset of $E^n$, which is also the solution set of a finite number of linear equations and nonlinear inequalities. Each iteration of the EA generates a PL-mapping (the convex polytopes of some covering of the domain into polytopes of some covering of the range, i.e., hyperplanes). These iterations do not stop until the termination criterion $\iota : \cup_{i=1}^{\infty} (H^u)^{(i)} \rightarrow$ {true, false} is met. If the termination criterion is met, the EA more than likely has found the most extreme translate of the hyperplane that passes through the polyhedron.

To define the termination criteria and the other EOs in more detail, we use the work of Lamont and Merkle [6]. In their work there is an ingenious use of random functions to generalize EA operators. They state that EA functions are mappings from parameter spaces to random functions. In order for us to use their work in this paper, we need to introduce a Borel algebra to our Euclidean topological space. This is easy to do because by definition the smallest $\sigma$-algebra containing all the open sets of a Euclidean topological space is a Borel algebra. And a Borel algebra has the necessary morphisms to allow our use of PL-mappings; also, a $\sigma$-algebra is bijective, an important property demanded of *c*.

Moreover, a Borel algebra over the set of real numbers defines a Borel measure. Given a real random variable defined on a probability space, its probability distribution is by definition also a measure of the Borel algebra[b].

To return to the work of Lamont and Merkle [6], we now define the EA fitness function:

Since *H* is a nonempty set, $c: A^l \rightarrow H$, and $f: H \rightarrow \mathbb{R}$, the fitness scaling function can be defined as $T_s: \mathbb{R} \rightarrow \mathbb{R}$ and a related fitness function as $\Phi \triangleq T_s \circ f \circ c$. In this definition it is understood that the objective function *f* is determined by the application, while the specification of the decoding function *c*[c] and the fitness scaling function $T_s$ are design issues.

Execution of an EA typically begins by randomly sampling with replacement from $A^l$. The resulting collection is the initial population, denoted as *P*. More generally, a population is a collection $P = \{a_1,...,a_\mu\}$ of individuals $a_i \in A^l$. Again, the number of individuals ($\mu$) is referred to as the population size.

Following initialization, execution proceeds iteratively. Each iteration consists of an application of one or more EOs. The combined effect of the EOs applied in a particular generation $t \in N$ is to transform the current population *P(t)* into a new population *P(t+1)*.

In the population transformation $\mu, \mu' \in \mathbb{Z}^+$ (the parent and offspring population sizes, respectively). A mapping $T : H^\mu \to H^{\mu'}$ is called a PT. If $T(P) = P'$, then *P* is a parent population and *P'* is the offspring population. If $\mu = \mu'$, then it is called simply the population size.

The PT resulting from an EO often depends on the outcome of a random experiment. In Lamont and Merkle [6] this result is referred to as a random population transformation (RPT or random PT). To define RPT, let $\mu \in \mathbb{Z}^+$ and $\Omega$ be a set (the sample space). A random function $R : \Omega \to T(H^\mu, \bigcup_{\mu' \in \mathbb{Z}^+} H^{\mu'})$ is called an RPT. The distribution of PTs resulting from the application of an EO depends on the operator parameters; in other words, an EO maps its parameters to an RPT.

Now that both the fitness function and RPT have been defined, the EO can be defined in general: let $\mu \in \mathbb{Z}^+$, *X* be a set (the parameter space) and $\Omega$ a set. The mapping

$$Z : X \to T\left(\Omega, T\left[H^\mu, \bigcup_{\mu' \in \mathbb{Z}^+} H^{\mu'}\right]\right)$$

is an EO. The set of EOs is denoted as $EVOP\ H, \mu, X, \Omega$.

There are three common EOs: recombination, mutation, and selection. These three operators are roughly analogous to their similarly named counterparts in genetics. The application of them in EAs is strictly Darwin-like in nature, i.e., "survival of the fittest."

In defining the recombination operator Lamont and Merkle [6] let $r \in EVOP\ H, \mu, X, \Omega$. If there exist $P \in H^\mu, \Theta \in X$ and $\omega \in \Omega$, such that one individual in the offspring population $r_\Theta\ P$ depends on more than one individual of *P*, then *r* is referred to as a recombination operator.

A mutation is defined in the following manner: let $m \in EVOP\ H, \mu, X, \Omega$. If for every $P \in H^\mu$, for every $\Theta \in X$, for every $\omega \in \Omega$, and if each individual in the offspring population $m_\Theta\ P$ depends on at most one individual of *P*, then *m* is called a mutation operator.

Finally, for a selection operator: let $s \in EVOP(H, \mu, X \times T(H, \mathbb{R}), \Omega)$. If $P \in H^\mu$, $\Theta \in X$, $\Phi: H \to \mathbb{R}$ in all cases, and *s* satisfies $a \in s_{\Theta,\Phi}(P) \Rightarrow a \in P$, then *s* is a selection operator.

**Unique triangulations and EAs**

Since EAs can generate convex hulls (the facets of the feasible set) and have been shown to be simplex-wise linear embeddings of the triangulation into Euclidean space, we use the work of Alboul and van Damme [7] and others to show that EAs can find the unique convex hull. Alboul and van Damme [7] consider an objective function that is a discrete measure of the *L1*-norm of the Gaussian curvature over a triangle mesh. This function has a very important property. As proven by Alboul and van Damme [7], the use of a local edge-swapping algorithm leads to the problem's unique global minimum, which corresponds to the unique convex hull. As noted by Effenberger [8], this measure (referred to in the literature as *tightness*) is a topological condition, meaning that any simplex-wise linear embedding of the triangulation into Euclidean space is as convex as possible. The measure can be understood as a generalization of the concept of convexity. Effenberger [8] proves that with regard to PL-embeddings of PL- manifolds, the tightness of combinatorial manifolds can be defined in a purely combinatorial way:

(i) A simplicial complex *K* that has a topological manifold as its underlying set *|K|* is called a triangulated manifold. *K* is called a combinatorial manifold of dimension *d* if all vertex links of *K* are PL *(d−1)* spheres, where a PL *(d −1)* sphere is a triangulation of the *(d −1)* sphere that carries a standard PL structure.

(ii) Let *G* be a field. A combinatorial manifold *K* on *n* vertices is called *(k-1)* tight with respect to *G* if its canonical embedding $K \subset \Delta^{n-1} E^{n-1}$ is *(k-1)* tight with respect to *G*, where $\Delta^{n-1}$ denotes the *(n−1)*-dimensional simplex.

Prestifilippo and Sprave [9] as well as Weinert *et al.* [10] have shown that EAs can replicate Alboul and van Damme's [7] edge-swapping algorithm; therefore, EAs can find the unique convex hull.

**Conclusions**

In this paper we show that PL-topology can explain the actions of EAs as they apply to solving convex optimization problems. We do so by establishing equivalence classes and their related quotient spaces. With these two properties and stellar subdivision we show EAs can create the convex hull built on vertices.

We introduce a Borel algebra to our Euclidean topological space to bring in the important work of Lamont and Merkle [6] on EA operators, e.g., mutation rate and selection.

Finally, with the work of Alboul and van Damme [7] and others, we show that EAs can indeed find the unique convex hull and are able to do so because of their PL-topology properties.

We have said nothing about nonconvex problems, but we note with some interest a statement at the very end of Alboul and van Damme [7]. They write that the use of EAs may give the necessary boost to the edge-swapping algorithm in the nonconvex case, i.e., EAs may be a way for the algorithm to avoid getting stuck in a local minimum. In this author's limited experience EAs can indeed solve nonconvex problems and can produce the same or better results as those generated by deterministic tools such as mixed-integer programming [11].

*References*

1. Sanderson, CP, Rourke, BJ: Introduction to Piecewise Linear Topology. Springer, New York (1972)

2. Hudson, JFP: Piecewise Linear Topology. WA Benjamin, New York (1969)

3. Radcliffe, NJ: Equivalence class analysis of genetic algorithms. Complex Systems, **5**, 183-205 (1992)

4. Radcliffe, NJ, Surry PS: Formal algorithms + formal representations = search strategies. In: Parallel Problem Solving in Nature IV, LNCS, pp. 366-375. Springer, Berlin (1996)

5. Coello, CA, Van Veldhuizen, D, Lamont, GB: Evolutionary Algorithms for Solving Multiobjective Problems. Kluwer Academic, New York (2002)

6. Lamont, LD, Merkle, JD: A random-based framework for evolutionary algorithms. In: Back, T. (auth.) Proceedings of the Seventh International Conference on Genetic Algorithms, pp. 105-112. Morgan Kauffamn, East Lansing, MI (1997)

7. Alboul, L, van Damme, R: Polyhedral metrics in surface reconstruction: tight triangulations. In: Martin T, Goodman R (eds.) The Mathematics of Surfaces VII, 1997. pp. 309-336. Midsomer Norton: The Institute of Mathematics and Its Applications, Dundee, Scotland (1997)

8. Effenberger, F: Stacked polytopes and the tight triangulation of manifolds. Journal of Combinatorial Theory, Series A. **18**, 6 (2011)

9. Prestifilippo, G, Sprave, J: Optimal triangulation by means of evolutionary algorithms. In: Conference Proceedings: Genetic Algorithms in Engineering. pp. 492-497. IEEE, Glasgow (1997)

10. Weinert, K, Menhen, J, Albersmann, F, Drerup, P: New solutions for surface reconstruction from discrete point data by means of computational intelligence. In: MJ and Martin RR Wilson (auths.) Mathematics of Surfaces: 10th IMA International Conference, pp. 431-438. Springer, New York (2003)

12. Clark, A, Kenyon, J: Using MOEAs to outperform stock benchmarks in the presence of typical investment constraints. Journal of Investing. **21**, 60-67 (2012)

---

[a] A simplex is defined as the convex hull of a set of *n+1* points, where *n* is the number of variables.

[b] This last statement is based on Von Neumann's theorem: every Borel set admits a section that is measurable with respect to all probability measures.

[c] Remember that if the domain of *c* is total, i.e., the domain of *c* is all of $A^{I}$, *c* is called a decoding function. The mapping of *c* is not necessarily surjective. The range of *c* determines the subset of $A^{I}$ available for exploration by the evolutionary algorithm.